\documentclass[fleqn,10pt]{wlscirep}
\usepackage[utf8]{inputenc}
\usepackage[T1]{fontenc}
\usepackage{soul}     
\title{ADMEDTAGGER: an annotation framework for distillation of expert knowledge for the Polish medical language}

\author[1,*]{Franciszek Górski}
\author[1]{Andrzej Czy{{\. z}}ewski}
\affil[1]{Gdansk University of Technology, Multimedia Systems Department, Gdansk, 80-233, Poland}

\affil[*]{franciszek.gorski@pg.edu.pl}


\begin{abstract}
In this work, we present an annotation framework that demonstrates how a multilingual LLM pretrained on a large corpus can be used as a teacher model to distill the expert knowledge needed for tagging medical texts in Polish. This work is part of a larger project called ADMEDVOICE, within which we collected an extensive corpus of medical texts representing five clinical categories - Radiology, Oncology, Cardiology, Hypertension, and Pathology. Using this data, we had to develop a multi-class classifier, but the fundamental problem turned out to be the lack of resources for annotating an adequate number of texts. Therefore, in our solution, we used the multilingual Llama3.1 model to annotate an extensive corpus of medical texts in Polish. Using our limited annotation resources, we verified only a portion of these labels, creating a test set from them. The data annotated in this way were then used for training and validation of 3 different types of classifiers based on the BERT architecture - the distilled DistilBERT model, BioBERT fine-tuned on medical data, and HerBERT fine-tuned on the Polish language corpus. Among the models we trained, the DistilBERT model achieved the best results, reaching an F1 score > 0.80 for each clinical category and an F1 score > 0.93 for 3 of them. In this way, we obtained a series of highly effective classifiers that represent an alternative to large language models, due to their nearly 500 times smaller size, 300 times lower GPU VRAM consumption, and several hundred times faster inference.
\end{abstract}
\begin{document}

\flushbottom
\maketitle
%
%
\thispagestyle{empty}

\twocolumn

\section{Introduction}
\label{sec:introduction}
\begin{figure}
  \centering
  \includegraphics[width=\columnwidth]{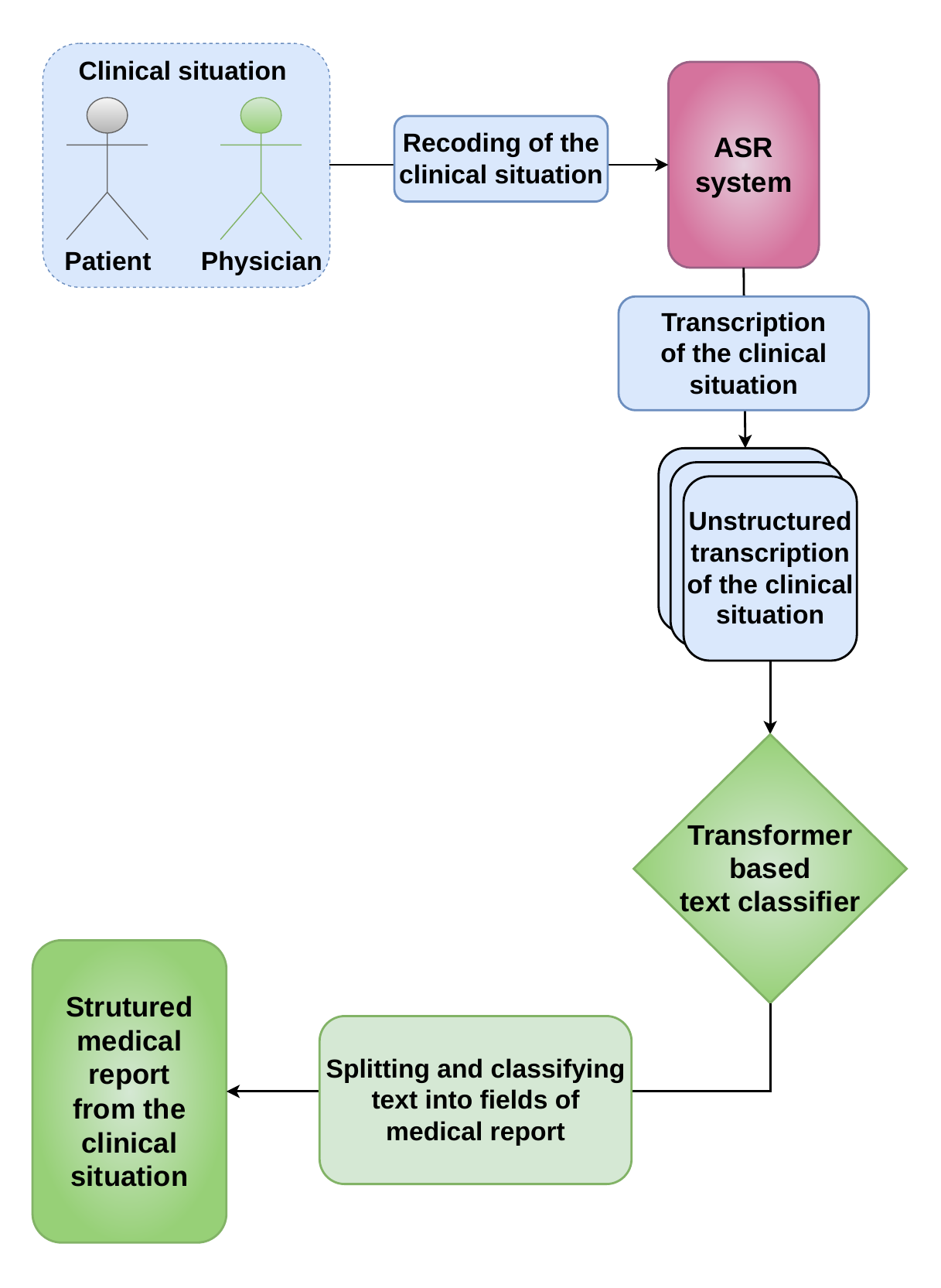} 
  \caption {Diagram of the ADMEDVOICE system.}
  \label{fig:admed_diagram}
\end{figure}

ADMEDVOICE is an R\&D project for the polish healthcare system which aim is to develop and implement a solution through which doctors can fill out medical records during the course of a medical interview, create descriptions (e.g., radiological) as needed, and prescribe treatment. The use case of filling out of medical records can be seen in Figure \ref{fig:admed_diagram}. The system to be developed will automatically generate templates for medical interviews and descriptions of available diagnostic results. In Polish healthcare institutions, doctors are typically not assisted by medical secretaries or medical assistants; therefore, the responsibility for preparing descriptions of examinations, visits, or hospitalizations rests solely on the physician, which consumes a significant amount of time. As a result, doctors have less opportunity to devote adequate attention to their patients. Physicians are aware of this issue, yet are unable to avoid the growing demands of medical documentation, they spend an increasing portion of their day at the computer. There is a clear need to explore ways to support healthcare through modern technologies that leverage information systems, especially by leveraging the latest advancements in artificial intelligence.

In this paper, we address the problem of insufficient labeled data for classifying specialized medical data. This is a prevalent issue in machine learning applications. Data labeling not only requires time and appropriate tools, but also often qualified experts who would handle the labeling process. Unfortunately, the work of these experts is often costly, and their time is limited. As a result, machine learning teams are unable to obtain the number of labels to conduct supervised training of a given machine learning model.

Therefore, the knowledge distillation approach is increasingly being used, in which a larger AI model - the teacher is employed to train a smaller model - the student. Knowledge distillation (KD), first formalized by Hinton et al. \cite{hinton2015distilling}, has evolved into a cornerstone technique for model compression and knowledge transfer. Several works and surveys have mapped this evolving landscape \cite{mansourian2025comprehensive, moslemi2024survey, acm2024survey, zhang2025cbkd, hemmatian2024uncertainty, hernandez2025kdintegrated}. Gou et al. \cite{gou2021knowledge} provide systematic review, categorizing KD methods from perspectives of knowledge categories, training schemes, teacher-student architectures, and distillation algorithms. KD also has application for LLMs. Xu et al. \cite{xu2024survey} presented a comprehensive survey specifically addressing KD's role in the LLM era, structured around three foundational pillars: algorithm, skill, and verticalization. Their work highlights KD's dual function: transferring advanced capabilities from proprietary models like GPT-4 to open-source alternatives (e.g., LLaMA, Mistral), and enabling model compression and self-improvement in open-source LLMs.

In this paper, we present a use case for medical data classification, in which a large language model (LLM) serves as the teacher model and a model from the BERT family as the student model.

During developing this publication, we were looking for answers to the following research questions:\newline
\textbf{RQ1}: Can a large language model (LLM), pre-trained on a vast corpus of textual data covering many thematic categories, effectively distill specialized expert knowledge from a specific domain using only a few textual examples?\newline
\textbf{RQ2}: Can a large language model effectively distill specialized medical knowledge in Polish, a language that is substantially less resourced and morphologically more complex than English?\newline
\textbf{RQ3}: Can a student model from the BERT family successfully learn a multiclass classification task on Polish medical texts using only labeled data produced through knowledge distillation?\newline

In the following paper we present ADMEDTAGGER - the annotation framework for knowledge distillation from Large Language Models to smaller, BERT-based models for the task oexpert validation used
to create evaluation setsf classification of Polish medical texts from different clinical categories.

ADMEDTAGGER is a response to real clinical needs requiring the development of a system for annotating electronic patient records into structured clinical forms. The learned annotations correspond to individual records in the forms, allowing for the quick generation of reports on the course of specific clinical situations in an automatic manner, thus freeing physicians from the very time-consuming procedure of preparing patient care notes and allowing them to devote significantly more attention to the patient. Our solution includes prompt structures for the teacher model, annotated text segmentation, training set generation, expert validation used to create evaluation sets, as well as train multiple students models. All of this comprises a framework for scalable annotation and expert knowledge distillation, used in this work to train medical text classifiers. ADMEDTAGGER is part of the ADMEDVOICE system, being prepared for implementation in Polish medical facilities.

\section{Related works}
\label{sec:related_works}
The introduction of BERT (Bidirectional Encoder Representations from Transformers) by Devlin et al. \cite{devlin2019bert} revolutionized text classification. Fields et al. \cite{fields2024survey} provided a comprehensive survey examining text classification with transformers. Their analysis reveals that while BERT-based models with 110-340M parameters established numerous benchmarks, more recent LLMs like GPT-4 have extended capabilities, particularly in handling longer contexts (up to 8,192 tokens compared to BERT's 512). Several studies have investigated specific aspects of BERT for text classification \cite{gasmi2022improving, patel2024survey, alnatsheh2024rethinking, sun2019finetune, sherstinsky2025bertattention, elmaaroufi2024lnlfbert}. Jiang et al. \cite{jiang2025progress} conducted a critical review asking "Are We Really Making Much Progress in Text Classification?" Their analysis revealed that fine-tuned discriminative models (particularly encoder-only transformers like BERT, RoBERTa, and DeBERTa) still define the state-of-the-art for single-label and multi-label text classification. Surprisingly, despite recent advances in LLMs, methods based on in-context learning with generative models do not generally outperform fine-tuned smaller language models (SLMs) in classification tasks. Even with advanced prompting techniques and ensemble methods, LLMs only marginally outperform encoder-only SLMs on individual datasets, requiring ensemble approaches to match or slightly exceed SLM performance.

The biomedical domain has seen substantial development of domain-specific language models \cite{lee2020biobert, rohanian2023lightweight, huang2020clinicalbert, peng2021bluebert, rasmy2021medbert, clinragen2025, kim2020attention, putelli2022bertmedical}. Lee et al. \cite{lee2020biobert} introduced BioBERT, pre-trained on PubMed abstracts and PMC full-text articles, which achieved state-of-the-art results on biomedical named entity recognition, relation extraction, and question answering. BioBERT's success demonstrated that domain-specific pre-training significantly improves performance on specialized tasks compared to general-domain BERT. Alsentzer et al. \cite{alsentzer2019publicly} released ClinicalBERT, specifically trained on clinical notes from the MIMIC-III database. Their work presented multiple variants: models initialized from BERT-base or BioBERT, and trained on all notes or only discharge summaries. Bio+Clinical BERT (initialized from BioBERT and trained on all MIMIC notes) showed particular promise for clinical NLP tasks, though interestingly performed worse on de-identification tasks—a consequence of differences between de-identified source text and synthetically non-de-identified task text. Wang et al. \cite{wang2020comparison} compared BERT implementations (vanilla BERT, BioBERT, and ClinicalBERT) for identifying complex medical concepts in narrative documents. Their results on three linguistically complex tasks (bariatric surgery discussions, statin non-acceptance, and tobacco use documentation) revealed that domain-specific models (BioBERT and ClinicalBERT) achieved superior performance overall, though neither consistently outperformed the other across all tasks. There is also a set of papers on employing LLMs in the biomedical domain like \cite{tanner2024llmdiagnostic, singhal2023medpalm, wang2025llmdiagnosis, liu2025llmmedical}.

Sakai and Lam \cite{sakai2025kdh} introduced KDH-MLTC (Knowledge Distillation for Healthcare Multi-Label Text Classification), a comprehensive framework leveraging model compression and LLMs specifically for medical text. Their approach transfers knowledge from BERT to DistilBERT through sequential training adapted for multi-label classification, with hyperparameter optimization via Particle Swarm Optimization (PSO). Experiments on the Hallmarks of Cancer (HoC) dataset achieved an F1 score of 82.70\% on the largest dataset variant, demonstrating that knowledge distillation can maintain high accuracy while significantly reducing model size and inference time. Hasan et al. \cite{hasan2025medical}, referenced in the KDH-MLTC study, achieved strong results using DischargeBERT and COReBERT as teacher models with BERT-PKD as the student model, obtaining AUC scores ranging from 72.36\% to 86.95\% across multiple clinical prediction tasks. These works demonstrated the viability of specialized clinical BERT variants as teachers for distillation in medical contexts.

Cross-lingual text classification has emerged as a crucial technique for leveraging resources from high-resource languages to enable classification in low-resource languages. Multiple approaches have been proposed to address this challenge \cite{karamanolakis2020cross, cho2024dsg, yang2022cross, katsogiannis2023multilingual, laurer2023cross, joshi2024universal}. These works tested many languages like Japanese, German, French, Chinese or Spanish.

There are not so many works on Polish language especially for the medical domain. Mroczkowski et al. \cite{mroczkowski2021herbert} introduced HerBERT, an efficiently pre-trained transformer-based language model specifically for the Polish. HerBERT, trained using knowledge transfer from multilingual to monolingual models, achieved state-of-the-art results on the KLEJ (Kompleksowa Lista Ewaluacji Językowych) benchmark across multiple downstream tasks. HerBERT's architecture uses the BERT-base configuration with careful consideration of Polish linguistic characteristics, particularly its rich inflectional morphology. The model has become the foundation for Polish NLP research, though no medical-domain variant has been developed. The KLEJ (Kompleksowa Lista Ewaluacji Językowych) \cite{rybak2020klej} benchmark provides nine evaluation tasks for Polish language understanding, including various text classification challenges. The Wroclaw Corpus of Consumer Reviews Sentiment (WCCRS) \cite{11321/700} includes medical domain reviews but focuses on sentiment analysis rather than medical text classification. While these resources support general Polish NLP development, specialized medical text corpora remain largely unavailable due to data privacy restrictions and limited annotation efforts.

The most significant work on Polish medical text classification comes from researchers at Górnośląskie Centrum Medyczne \cite{polish2022deep}, who analyzed Polish electronic health records for diagnosis prediction in cardiovascular disease patients. Using 50,465 hospitalization records, they applied transformer language models (RoBERTa and BERT) trained on Polish or multilingual data for multi-label ICD-10 diagnosis classification. Their models achieved 70\% accuracy using only patient medical history (average 132 words) and 78\% accuracy when incorporating additional hospitalization information. This work demonstrated that state-of-the-art NLP techniques can effectively process Polish clinical text, though the absence of Polish-specific medical language models limited performance compared to English medical NLP systems. The AssistMED project \cite{assistmed2024}, documented in Polish Archives of Internal Medicine, developed NLP tools for observational clinical research data retrieval from Polish electronic health records. Their practical use case highlighted several challenges: limited availability of Polish medical terminologies, insufficient NLP tools designed explicitly for Polish clinical text, and disappointing performance of general-purpose LLMs like ChatGPT on Polish medical licensing examinations. The project relied primarily on dictionary-based and rule-based approaches rather than deep learning, reflecting the scarcity of annotated Polish medical corpora. Zielonka et al. \cite{zielonka2025machine} presents work in which they evaluate LLMs in the task of categorizing medical information into categories such as medications, diseases, symptoms, or procedures and compare them with annotations made by medical experts, demonstrating high concordance of the GPT-4 model with expert responses. Czyżewski et al. \cite{czyzewski2025comprehensive} presents work in which a broad corpus of medical speech recordings and transcriptions from physicians is described and made available, containing names of medications and diseases spoken by specialists from many clinical fields. This corpus contains nearly 15 hours of recordings from 28 different speakers, and was additionally expanded synthetically with an additional 83 hours of recordings. Such a corpus constitutes a potential data source that can be used for training language models. 

Several critical gaps emerge: 1) Absence of Polish medical language models, 2) limited Knowledge Distillation for tasks for Polish medical language, 3) scarcity of annotated Polish medical corpora and 4) cross-lingual transfer potential unexplored.

In our work we demonstrate the framework for effective Knowledge Distillation for the classification of the Polish medical texts, gathering a wide corpora of Polish medical texts for many clinical categories, exploring potential  for cross-lingual transfer resulting in developing Polish medical language models for classification tasks.

\section{ADMEDTAGGER}
\label{sec:methodology}
In our work, we encountered the problem of having access to a relatively large amount of data but lacking sufficient labels for the data classes. Labeling this data required specialists in the form of physicians from specific medical domains. However, their time was severely limited, and they were unable to label a sufficient amount of data. Therefore, we needed a solution that would allow us to develop a series of multiclass classifiers with a limited number of labeled samples.

\begin{figure*}
  \centering
  \includegraphics[width=.85\textwidth]{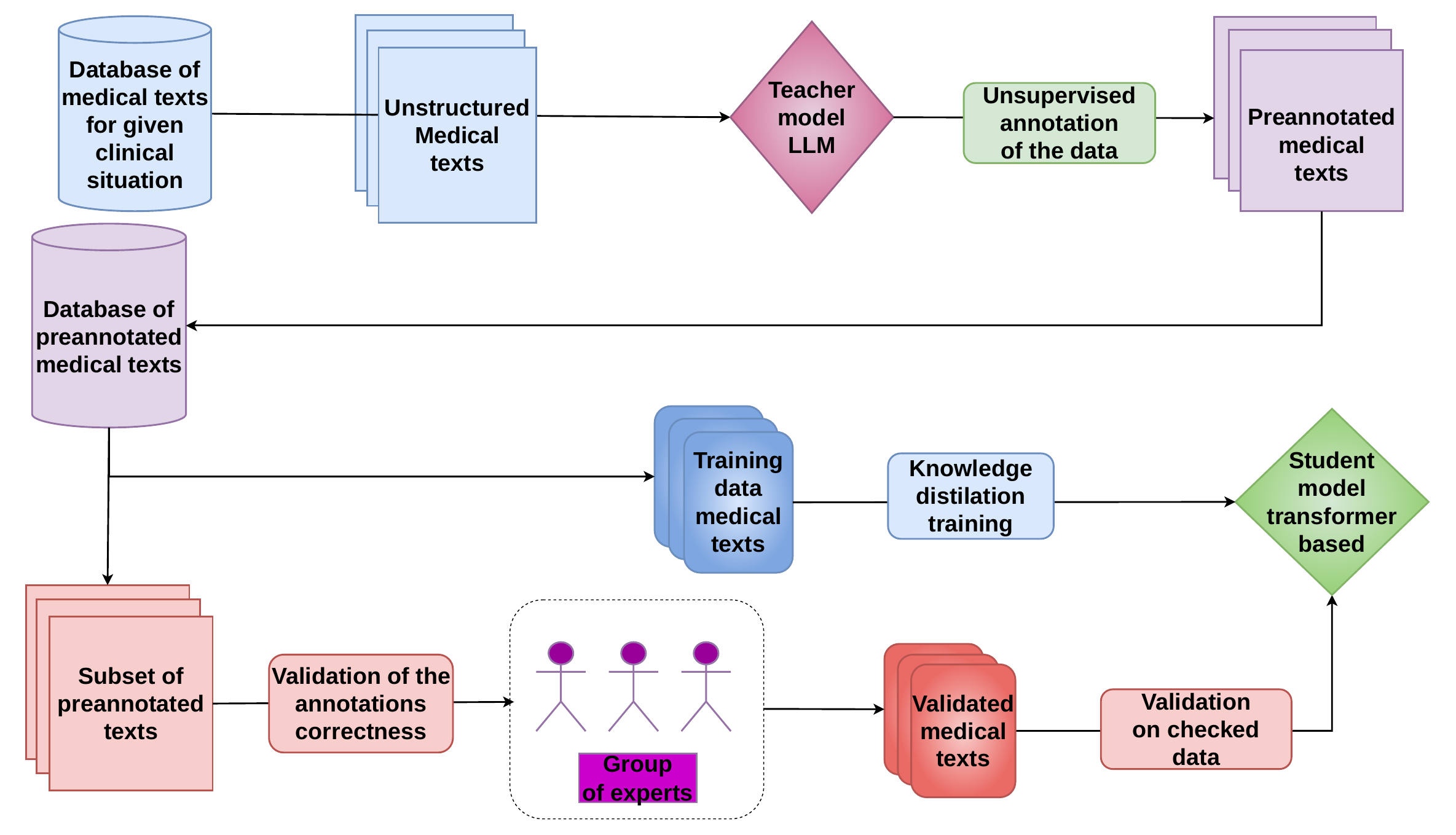} 
  \caption {Diagram of the whole ADMEDTAGGER methodology.}
  \label{fig:admettagger_diagram}
\end{figure*}

In this paper, we present the solution whose diagram is shown in Figure \ref{fig:admettagger_diagram}. In our approach, we used the physicians' limited time to label only small subsets of data from each specialization, then split these into a test set and a few contextual examples. The contextual examples were subsequently inserted into the prompt of a large language model to enable few-shot contextual learning. The LLM was used as the teacher model for labeling the remaining data in each specialization. For this task, we selected the Llama3.1-70B model. Unlike classical pseudo-labeling, the teacher model transfers structured, domain-specific annotation behavior derived from expert-designed clinical forms and ontologies.

\subsection{Teacher and Student Models}
The knowledge distillation approach relies on the operation of two models: a teacher model and a student model. In our solution, we chose to use a relatively large LLM as the teacher model—one pretrained on an extensive data corpus that potentially includes not only many natural languages other than English, but also content from various thematic domains, including medical texts. Therefore, we selected the Llama3.1-70B-Instruct model, a multilingual large model pretrained on a corpus of up to 15 trillion tokens. The model is available as open weights and can be used locally. We used the Instruct version, which is adapted for handling instruction-based prompting.

As the student model, we selected three models from the BERT family. The chosen models were: 1) HerBERT (HerBERT-base-cased) - explicitly trained on a Polish language corpus, 2) BioBERT (BioBERT-base-cased-v1.1) - trained on a corpus of English medical texts, and 3) DistilBERT (DistilBERT-base-multilingual-cased) - a specially distilled multilingual model. All of these models contain no more than 134 million parameters. This selection allowed us to achieve potentially very large computational cost reductions, shrinking the model size by more than 500 times while narrowing its functionality to the task of multiclass classification of medical texts.

\subsection{Prompt Construction}
\begin{figure}
  \centering
  \includegraphics[width=\columnwidth]{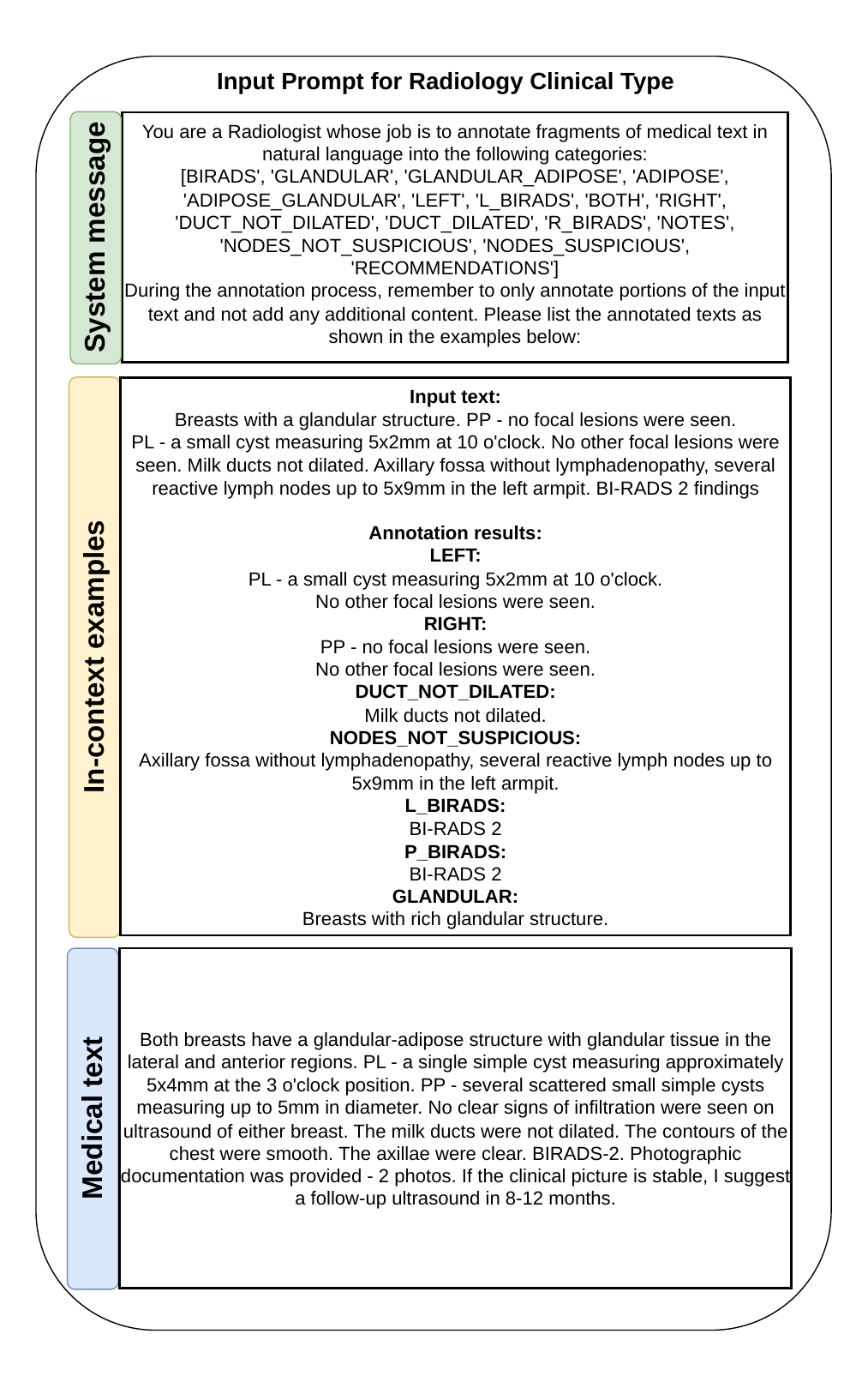} 
  \caption {Prompt construction for the Radiology Clinical Type.}
  \label{fig:prompt}
\end{figure}

The structure of the prompt used for every clinical situation is shown in Figure \ref{fig:prompt}. It shows the example for a radiology situation, but for every other clinical situation the structure remains with changes only to the list of labels and In-context examples.
The prompt consists of:\newline
\textbf{System message}: for each specialization, it contains a brief description indicating which medical specialty the model should assume and what labels/classes are available for that specialization.\newline
\textbf{In-context examples}: demonstrating how the text should be classified into the respective classes and what the expected output format should be.\newline
\textbf{Medical text}: input text that needs to be classified.

\subsection{Datasets}
\begin{figure*}
  \centering
  \includegraphics[width=\textwidth]{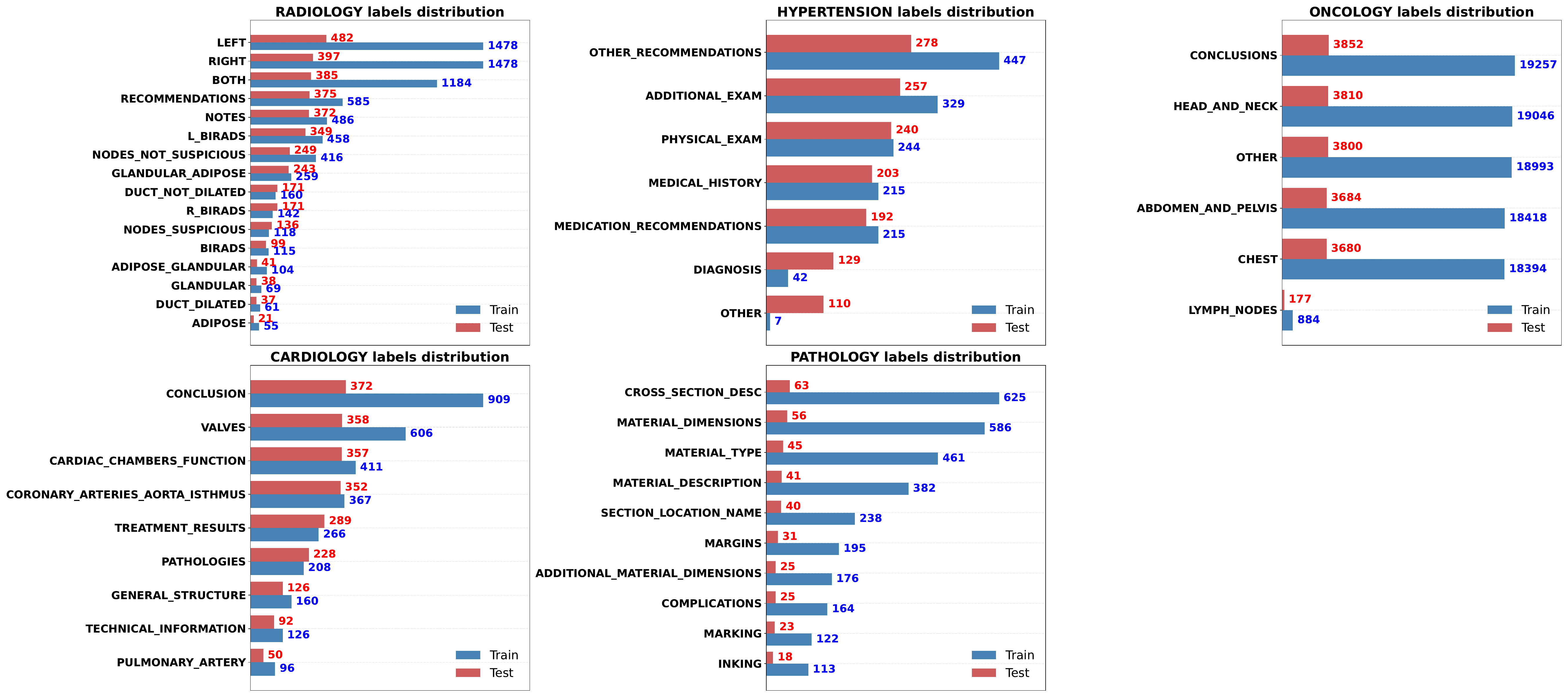} 
  \caption {Datasets' labels distribution for each clinical category.}
  \label{fig:datasets_distribution}
\end{figure*}

In our work, we used medical datasets in Polish. These datasets have been collected as part of the ADMEDVOICE project. The detailed description regarding data collection procesess has been provided in the following works: Zielonka et. al. \cite{zielonka2025machine} and Czyżewski et. al. \cite{czyzewski2025comprehensive}.
As part of the project, we collected a total of 143,756 medical texts from five clinical scenarios: Cardiology, Radiology, Oncology, Pathomorphology, and Hypertension but limited each category to 5000 samples which gives us 21,073 texts, because Pathology and Hypertension did not reach 5000. As a result, we obtained a substantial set of real clinical texts representing important areas of medicine. Each clinical scenario dataset has a separate set of tags that denote specific types of information relevant to documenting the course of that clinical case and tailored to the practical needs of the participating clinicians. During the knowledge distillation stage, the teacher model annotates entire texts with segments corresponding to individual concepts/tags from the forms associated with each clinical scenario. Since the task designed for the student models is multiclass text classification, the full texts were divided during the distillation stage into segments that correspond to a given tag/section of the clinical form. The detailed distribution of tags in the training and test sets is shown in the charts in Figure \ref{fig:datasets_distribution}. Test samples for each category have been chosen randomly while ensuring the presence of samples for each of the labels. This scenario was dictated by the limited availability of the medical experts collaborating with us, due to which we were unable to conduct validation on a large dataset. To our knowledge, the risk of overlap between distributional or procedural effects that could inflate performance estimates was minimal and did not significantly affect the results of our experiments. These charts clearly show that in 4 out of the 5 training datasets there is a pronounced imbalance in the representation of specific tags. However, this class imbalance problem was addressed during training by assigning weights to each class when training the classification model. In Table \ref{tab:data_examples} we present the example of the input text, predicted label ID and name of the predicted label by BERT model for each of the evaluated clinical categories. From this example you can clearly seen that the task for BERT models is prediction of the class ID, which represents the type of clinical information important for the physician, for the given text sequence.

\begin{table*}
\centering
  \resizebox{\textwidth}{!}{
        \large
        \begin{tabular}{|c|p{10cm}|c|c|}
        \hline
        \textbf{Clinical category} & \textbf{Input text} & \textbf{Predicted Label ID} & \textbf{Predicted Label} \\
        \hline
        Cardiology & Atrioventricular and ventriculo-vascular connections are concordant. Venous drainage to the atria is normal. There is no pericardial effusion. & 0 & GENERAL\_STRUCTURE \\
        \hline
        Oncology & Glucose metabolism image normal. Right thyroid lobe enlarged. & 1 & HEAD\_AND\_NECK \\
        \hline
        Pathology & In plastic compartments and loosely in the container, several breast gland biopsies were submitted. & 1 & MATERIAL\_DESCRIPTION \\
        \hline
        Radiology & Milk ducts bilaterally not dilated. & 4 & DUCT\_NOT\_DILATED \\
        \hline
        Hypertension & The patient with suspected sleep apnea was admitted to the Arterial Hypertension Clinic for a sleep study. & 0 & MEDICAL\_HISTORY \\
        \hline
        \end{tabular}      
   }
\caption{Examples of input texts and outputs from the fine-tuned BERT models for each of the evaluated clinical category.}
\label{tab:data_examples}
\end{table*}

\subsection{Metrics}
In this study, we intentionally did not use offset-based annotation, because our objective was not to evaluate span-level extraction accuracy, but rather to assess whether the teacher LLM could correctly associate a text sequence with its corresponding clinical label. In other words, the focus of the study is on label assignment at the sequence level, not on the exact localization of textual spans. The evaluation is performed by comparing the predicted label with the ground truth label for the given input text to the BERT model. The BERT models are tasked with predicting the label for a given text sequence, not with extracting subtexts with associated labels.
We decided to evaluate results with the following metics:\newline
\textbf{F1 score} - is a harmonic mean of the precision and recall, where an F1 score reaches its best value at 1 and worst score at 0. The relative contribution of precision and recall to the F1 score are equal. We calculate this metric for each label, and find their unweighted mean.\newline
\textbf{AUROC} - Area Under the Receiver Operating Characteristic Curve (ROC AUC). he ROC curve was calculated using prediction scores for each test input. The metric was calculated for each label and then reported as an unweighted mean across all labels for a given clinical category. \newline
\textbf{Accuracy} - This accuracy refers to the accuracy of the labels predicted by the BERT model. We used the predicted labels for the test input texts and the ground truth labels associated with each text. It is simply the fraction of correct classifications with 95\% confidence interval. \newline
\textbf{Confidence intervals} - confidence intervals were computed from the results of the evaluation on the whole test set without repetition, using the Wilson score interval to compare the predictions of the BERT model with the ground truth labels.\newline

\section{Experiments}
\label{sec:experiments}
Our experiments followed the pipeline shown in Figure \ref{fig:admettagger_diagram}. First, the teacher model pre-annotated texts for each clinical scenario. From this annotated set, we extracted a subset containing representations of all tags for the given clinical scenario. This subset was then validated by medical specialists—physicians from the respective fields—who reviewed the annotations and introduced corrections where errors appeared. All annotators were provided with clear annotation guidelines, including definitions of the labels. Each clinical category was assigned 2 specialists who evaluated the annotations made by the LLM. Each instance was revised by 2 specialists – one made the initial revision, and then the second, more experienced specialist revised their decision. In case of disagreement, we always chose the revision of the second, more experienced specialist. The next step was training all three classifier models for each clinical scenario. Finally, each trained BERT classifier was evaluated on the physician-validated test sets. In our work, we did not address the problem of first splitting the entire input text into segments and classifying individual segments, focusing instead on the models' capabilities for classifying ready-made text fragments. To our knowledge, the problem of addressing these 2 tasks simultaneously — i.e., having the model operate on the entire text and divide it into individually classified segments — could be tackled using an approach of training a BERT model for a token classification task with a BIO (Begin/Inside/Outside) or BIOES tagging scheme. Each token receives a label such as B-LEFT, I-LEFT, B-RECOMMENDATIONS, O, etc. The model learns both where the boundaries are and what the class is. In this way, it would be possible to train an end-to-end model capable of performing both of these tasks entirely. We did not undertake this task in the present work, as this approach was not known to us at the time and we wished to focus primarily on analyzing the capabilities of BERT models for training on distilled labels, without the need to burden the training process with an additional task related to segmenting the text into fragments.

\subsection{Experimental Settings}
All experiments were conducted on a computing server purchased as part of the ADMEDVOICE project – Adaptive Intelligent Speech Processing System for Medical Personnel with the Structuring of Test Results and Support of the Therapeutic Process, using 4 NVIDIA L40 GPUs, each with 48GB of VRAM, and an AMD EPYC 75F3 32-Core Processor. The LLama3.1-70B-Instruct model required 140GB of free disk space and the size of trained BERT models was 4.6GB per model, which results in 69GB of free disk space. All scripts were executed with Python 3.10. The pretrained LLM and BERT weights were utilized through Hugging Face’s \textit{transformers} library (version 4.33.1) and \textit{accelerate} (version 0.33.0), with \textit{CUDA} 12.1. The models were loaded using the \textit{AutoModelForCausalLM}, \textit{AutoModelForSequenceClassification} and \textit{AutoTokenizer} classes with the options \textit{device\_map="auto"} and \textit{torch\_dtype=torch.bfloat16}, enabling multi-GPU inference and reduced GPU memory usage for LLama model.

Training of classifiers was conducted on a single NVIDIA L40 GPU with batch\_size=512, grad\_acc\_steps=1, learning\_rate=1e-6 and num\_epochs=100.

\section{Evaluation results}
\label{sec:results}
\textbf{Model Performance Across Clinical Domains:}\newline
In Table \ref{tab:clf_results} we can see the results of DistilBERT, BioBERT, and HerBERT models across five clinical domains, with DistilBERT consistently outperforming the other models in each settings. DistilBERT achieves the highest F1 scores, AUROC values and accuracies with confidence intervals indicating stable and reliable performance. In these domains, BioBERT and HerBERT has lower results than DistilBERT, with BioBERT exhibiting the weakest results overall. HerBERT performs moderately, consistently surpassing BioBERT yet remaining below DistilBERT on all reported metrics.

In Cardiology, DistilBERT shows strong performance (F1 score = 0.97, AUROC = 0.98, Accuracy = 0.98), outperforming both BioBERT and HerBERT. Similar patterns are observed in Hypertension and Radiology, where DistilBERT has higher F1 scores and narrower confidence intervals, suggesting robust generalization. In Pathology, the gap between DistilBERT and the other models remains substantial, with DistilBERT achieving an F1 score of 0.93 compared to 0.83–0.88 for BioBERT and HerBERT. The Oncology domain represents an exception - all three models achieve near-perfect classification performance (F1 $\approx$ 0.99; AUROC $\approx$ 1.00; Accuracy $\approx$ 0.99). The nearly identical metrics and overlapping confidence intervals indicate that the models behave equivalently in this domain.

Overall, these results demonstrate that DistilBERT provides the most reliable and accurate performance in all clinical domains, while BioBERT underperforms consistently. The results in Oncology suggests the presence of an easier classification landscape or highly distinctive class boundaries, leading all models to converge toward similar near-optimal decisions.

Additionally, Figure \ref{fig:confusion_matrices} shows the confusion matrices obtained for the best-performing model - DistilBERT, for each clinical scenario.
\newline\textbf{Interpretation of 95\% Confidence Intervals:}\newline
The 95\% confidence intervals for accuracy, reported in the Table \ref{tab:clf_results}, provide important insight into the reliability and stability of the models’ performance across clinical domains. In all domains except Oncology, DistilBERT not only achieves the highest point estimates of accuracy but also shows consistently narrow confidence intervals, indicating low variability and high certainty in its performance. For example, in Cardiology the interval spans only approximately ±0.006 from the estimated accuracy (0.9730–0.9848), whereas the corresponding intervals for BioBERT and HerBERT are wider. This pattern suggests that DistilBERT not only performs better on average but does so with greater consistency across samples. A similar trend is observed in Hypertension, Radiology, and Pathology, where DistilBERT maintains narrower intervals compared to both BioBERT and HerBERT. The comparatively broader intervals for BioBERT, particularly in domains where its performance is weak (e.g., Hypertension: 0.4892–0.5413), indicate greater variability and lower confidence in the stability of the model’s predictions. HerBERT typically falls between two other models - is more stable than BioBERT but still shows wider intervals than DistilBERT. The Oncology domain again is an exception. All three models achieve nearly identical accuracies with extremely narrow confidence intervals (e.g., ~0.9922–0.9946), reflecting very high consistency in predictions. This tight clustering of intervals reinforces the conclusion that Oncology constitutes a comparatively easy classification scenario in which model performance is both uniformly high and highly stable.

Taken together, the confidence interval analysis strengthens the conclusions drawn from the point estimates: DistilBERT delivers not only superior accuracy but does so with greater statistical certainty across most clinical domains. The width of the intervals complements the hypothesis that BioBERT's lower performance is accompanied by greater instability, while HerBERT remains competitive yet consistently less robust than DistilBERT. The uniformly narrow intervals in Oncology confirm the domain’s distinct nature, in which all models converge toward highly reliable, near-optimal classification outcomes.
\newline\textbf{Models comparison with non-parametric tests:}\newline
Pairwise Wilcoxon signed-rank tests were applied to compare model performance across clinical domains. The results from the Table \ref{tab:wilcoxon_all_domains} show substantial variation in the adequate sample size used by the test (n\_effective), which reflects not the number of available samples but the number of instances where two models produce different correctness outcomes. This distinction is essential for interpreting the Oncology results.

In Cardiology, Radiology, Hypertension and Pathology, model predictions differed across samples, resulting in large n\_effective values (20–529). Consequently, the Wilcoxon test yielded highly significant differences between all model pairs (p-values from 10e-8 to 10e-79), indicating robust and systematic performance differences in these domains. In contrast, the Oncology dataset—despite being the largest in absolute size - produced very small effective sample sizes (n\_effective = 2–5). This indicates that the models made almost identical correct/incorrect decisions on nearly all Oncology samples. As a result, the Wilcoxon test has very limited power in this domain, with only the BioBERT vs. HerBERT comparison reaching significance (p = 0.02). The remaining non-significant results should therefore be interpreted not as evidence of equal performance, but as a consequence of the extremely small number of disagreement cases between models.

Overall, the statistical analysis reveals that meaningful model differences emerge clearly in domains where models disagree on a substantial proportion of samples. In domains where predictions are nearly identical, as observed in Oncology, the Wilcoxon test lacks the sensitivity to detect differences, even when the dataset itself is large.
\newline\textbf{Impact of Knowledge Distillation on Computational Efficiency:}\newline
In Table \ref{tab:vram_time}, we observe measurements for each of the 3 classifier models and the teacher model (Llama3.1) regarding GPU VRAM memory consumption and inference time for the same input text, which was divided into all labels (text field types) recognized within it. These measurements were performed for each of the 5 clinical situations. The measurement results showed enormous differences in both VRAM consumption and inference time between the Llama3.1 model and all BERT models. In the case of the Llama3.1 model, the demand for VRAM memory during inference was very high and ranged between 130 and 150 GB. Meanwhile, for BERT models, this consumption did not exceed 0.55 GB, and was the highest for the DistilBERT models. This results in approximately a 300-fold difference in VRAM consumption for classifying the same text. Similarly unfavorable differences in favor of the LLM occur in the case of inference time, although the Llama3.1 results are more varied here - the fastest inference took place for Pathology - just under 25 seconds, while the longest was in the case of Oncology, lasting over 113 seconds. For BERT models, inference time was consistently low and rarely exceeded 0.01 seconds. The exception here was the inference of the DistilBERT model for Cardiology.

These measurements demonstrate a clear advantage of BERT models over large language models such as Llama3.1-70B.

\begin{table*}
\centering
  \resizebox{\textwidth}{!}{
        \large
        \begin{tabular}{c|cccc|cccc|ccccc}
        \toprule
         \multicolumn{5}{c}{\textbf{DistilBERT}} & \multicolumn{4}{c}{\textbf{BioBERT}} & \multicolumn{4}{c}{\textbf{HerBERT}} \\
         \midrule
         Clinical Type & F1 score & AUROC & Accuracy & 95\% CI & F1 score & AUROC & Accuracy & 95\% CI & F1 score & AUROC & Accuracy & 95\% CI  \\
        \midrule
        Cardiology & \textbf{0.97} & 0.98 & 0.98 & (0.9730--0.9848) & 0.75 & 0.90 & 0.85 & (0.8358--0.8654) & 0.85 & 0.92 & 0.90 & (0.8908--0.9153) \\
        Hypertension & \textbf{0.81} & 0.90 & 0.82 & (0.8040--0.8437) & 0.49 & 0.78 & 0.52 & (0.4892--0.5413) & 0.53 & 0.79 & 0.56 & (0.5375--0.5892) \\
        Radiology & \textbf{0.87} & 0.93 & 0.90 & (0.8890--0.9088) & 0.75 & 0.87 & 0.78 & (0.7625--0.7899) & 0.79 & 0.88 & 0.82 & (0.8062--0.8314) \\
        Pathology & \textbf{0.93} & 0.97 & 0.94 & (0.9109--0.9601) & 0.83 & 0.91 & 0.85 & 0.8100--0.8830) & 0.88 & 0.94 & 0.89 & (0.8550--0.9189) \\
        Oncology & \textbf{0.99} & 1.00 & 0.99 & (0.9922--0.9945) & 0.99 & 1.00 & 0.99 & (0.9920--0.9943) & 0.99 & 1.00 & 0.99 & (0.9923--0.9946) \\
        \bottomrule
        \end{tabular}      
   }
\caption{Classification metrics of the trained BERT models on the data labeled by LLM}
\label{tab:clf_results}
\end{table*}

\subsection{Common errors}
\begin{figure*}
  \centering
  \includegraphics[width=\textwidth]{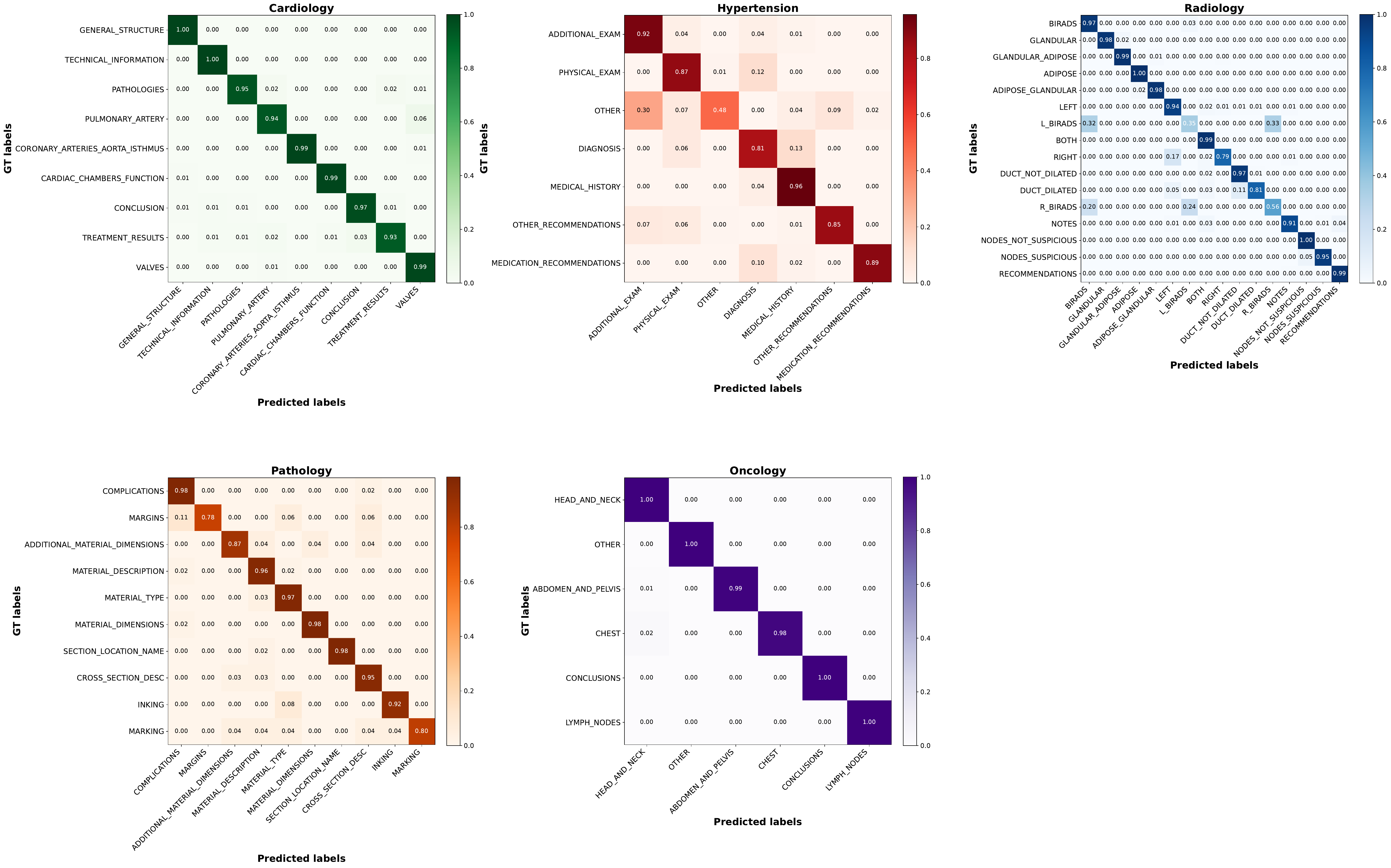} 
  \caption {DistilBERT's confusion matrices for each clinical category.}
  \label{fig:confusion_matrices}
\end{figure*}

The DistilBERT model, despite being our best performer, shows its highest error rate in the Hypertension category, where it achieves an F1 score of 0.81. From the Figure \ref{fig:confusion_matrices}, we see that the model struggles most with the OTHER label in this category, achieving only 48\% accuracy, mainly confusing it with ADDITIONAL\_EXAM label in 30\% of cases. While the exact cause isn't clear, this might stem from an imbalance in the training data that favors ADDITIONAL\_EXAM. The model also underperforms on the DIAGNOSIS label, with 81\% accuracy, frequently confusing it with MEDICAL\_HISTORY label in 13\% of cases. This particular confusion is understandable - both labels can legitimately apply to certain text fragments, and in this case the decision was made by a team of experts, doctors specializing in the given field. The clinical category with the second lowest F1 score is Radiology. Here, the DistilBERT model made the most mistakes for the L\_BIRADS (left) and R\_BIRADS (right) labels, achieving only 38\% and 56\% accuracy respectively. Both labels are confused with each other and with the BIRADS (both) label. Another common mistakes occur with RIGHT (breast) label and DUCT\_DILATED. The last clinical category which shows some significant labeling errors is Pathology, where DistilBERT model has some problems with MARGINS and MARKING labels, having 78\% and 80\% accuracy respectively. The label MARGINS has been mistaken with label COMPLICATIONS in 11\% of cases. From the analysis of the test input data we can see that once again these two labels will fit the text but the doctors choose the label MARGINS instead of MARKING.

\begin{table}
\centering
\resizebox{\columnwidth}{!}{
\begin{tabular}{llllll}
\toprule
Model\_1 & Model\_2 & stat & p\_value & n\_effective \\
\midrule
\multicolumn{5}{c}{Cardiology} \\
\midrule
BioBERT & DistilBERT & 480.0 & 3.43e-39 & 191 \\
BioBERT & HerBERT & 4065.5 & 2.72e-09 & 172 \\
DistilBERT & HerBERT & 768.0 & 6.25e-20 & 127 \\
\midrule
\multicolumn{5}{c}{Oncology} \\
\midrule
BioBERT & DistilBERT & 3.0 & 0.17 & 5 \\
BioBERT & HerBERT & 0.0000 & 0.02 & 5 \\
DistilBERT & HerBERT & 0.0 & 0.15 & 2 \\
\midrule
\multicolumn{5}{c}{Pathology} \\
\midrule
BioBERT & DistilBERT & 18.5 & 1.45e-08 & 36 \\
BioBERT & HerBERT & 246.0 & 0.01 & 40 \\
DistilBERT & HerBERT & 10.5 & 5.69e-05 & 20 \\
\midrule
\multicolumn{5}{c}{Radiology} \\
\midrule
BioBERT & DistilBERT & 15635.0 & 2.03e-71 & 529 \\
BioBERT & HerBERT & 42644.5 & 4.63e-11 & 492 \\
DistilBERT & HerBERT & 10830.0 & 3.39e-42 & 379 \\
\midrule
\multicolumn{5}{c}{Hypertension} \\
\midrule
BioBERT & DistilBERT & 1985.0 & 1.26e-79 & 396 \\
BioBERT & HerBERT & 8977.5 & 8.99e-54 & 398 \\
DistilBERT & HerBERT & 2866.5 & 1.82e-08 & 146 \\
\bottomrule
\end{tabular}
}
\caption{Pairwise Wilcoxon signed-rank test results for model accuracies across all clinical types.}
\label{tab:wilcoxon_all_domains}
\end{table}

\begin{table}
\centering
\resizebox{\columnwidth}{!}{
\begin{tabular}{lllll}
\toprule
\multicolumn{5}{c}{VRAM Usage [GB]} \\
\midrule
Clinical Type & Llama3.1 & DistilBERT & BioBERT & HerBERT \\
\midrule
Cardiology & 141.84 & 0.52 & 0.43 & 0.48 \\
Hypertension & 159.59 & 0.52 & 0.43 & 0.48 \\
Radiology & 141.71 & 0.55 & 0.46 & 0.50 \\
Pathology & 138.77 & 0.52 & 0.42 & 0.48 \\
Oncology & 136.53 & 0.55 & 0.46 & 0.50 \\
\midrule
\multicolumn{5}{c}{Inference time [s]} \\
\midrule
Clinical Type & Llama3.1 & DistilBERT & BioBERT & HerBERT \\
\midrule
Cardiology & 68.57 & 0.17 & 0.01 & 0.01 \\
Hypertension & 106.89 & 0.01 & 0.01 & 0.01 \\
Radiology & 66.44 & 0.01 & 0.01 & 0.01 \\
Pathology & 24.76 & 0.01 & 0.01 & 0.01 \\
Oncology & 113.30 & 0.01 & 0.01 & 0.01 \\
\bottomrule
\end{tabular}
}
\caption{VRAM Usage and Inference time comparison between Teacher LLM and students BERT models across Clinical domains.}
\label{tab:vram_time}
\end{table}

\section{Conclusions}
\label{sec:conclusions}
In our work, we presented one of the first knowledge distillation frameworks for medical text classification tasks in Polish under limited data labeling capabilities. Our method was tested across a wide range of medical domains, utilizing data from as many as 5 different fields - hypertension, cardiology, pathology, oncology, and radiology. We tested 3 different types of BERT models: 1) DistilBERT, a model that has already undergone general knowledge distillation, 2) BioBERT, a model fine-tuned on medical data in English, and 3) HerBERT, a model fine-tuned on Polish language data. Our results showed the DistilBERT model's unquestionable advantage in assimilating knowledge distilled by the Llama3.1 LLM for Polish medical text classification. It appears that the cause of these differences may be the pretraining stage of each of these models and the language corpus was used for this purpose. The DistilBERT model is multilingual, trained on a Wikipedia corpus containing as many as 104 languages, including Polish. The Polish language corpus is a dump of Wikipedia articles, and therefore, it undoubtedly also includes a collection of medical texts. Thus, during the pretraining phase, the model was exposed to medical data in Polish. The HerBERT model was pretrained on Polish language data, but devoid of any medical texts, which resulted in its weaker performance compared to the DistilBERT model. The BioBERT model typically achieves the worst results of the three, because it is an English-centric model, pretrained on medical data, but only in English, which may have created significant barriers in learning Polish medical texts.
In this way, we obtained a model that contains over 500 times fewer parameters, which translates to nearly 300 times lower GPU VRAM requirements and takes on average several hundred times less time for inference than the Llama3.1 teacher model, while achieving very high classification efficiency measures by F1 score, AUROC and Accuracy.

The inference parameters of the distilled classifiers we achieved, such as execution time <= 0.01 s, GPU VRAM consumption <= 0.55 GB, as well as their effectiveness, allow for the real implementation of our proposed system in medical facilities, for example, for tasks involving filling in electronic health records (EHR). The short inference time ensures the possibility of real-time operation, while the low computational requirements allow for inference on non-specialized computer hardware equipped with only a CPU instead of a GPU. Additionally, BERT models trained for classification tasks guarantee repeatability and stability of results. Our solution is an alternative to using an LLM model as an external service, ensuring "in-house" operation, solving the problem of data transfer to external services, which allows for compliance with regulations related to personal/patient data protection such as the Polish General Data Protection Regulation (GDPR). The framework we proposed also reduces costs resulting from payments associated with each use of such an external LLM service for every inference, which radically reduces the operating costs of the system at large scale.

Of course, in our solution there is a risk of bias propagation from the teacher models, the dependence of quality on the fit of the prompt used by the teacher, as well as the need for periodic expert validation of the distilled labels.

We show that a pre-trained multilingual large language model like Llama3.1 can effectively distill specialized expert knowledge from a specific domain like the medical domain. We show that a multilingual LLM like Llama3.1 can be efficiently apply for knowledge distillation in Polish language. Finally we show that BERT models, especially DistilBERT, successfully learn a multiclass classification task solely from data with distilled labels.

\section*{Declarations}

\section*{Funding}
The Polish National Centre for Research and Development (NCBR) supported this research in the project: “ADMEDVOICE- Adaptive intelligent speech processing system of medical personnel with the structuring of test results and support of therapeutic process,” no. INFOSTRATEG4/0003/2022.

\section*{Data availability}
The anonymized datasets generated during and analyzed during the current study are available from the corresponding author upon reasonable request.

\section*{Code availability}
The scripts with the implementation of all the algorithms used during the experiments are available in this GitHub repository: \url{https://github.com/Fragorsk/ADMEDTAGGER}.

\section*{Author contributions}
Conceptualization, F.G. and A.C.; Methodology, F.G.; Supervision, A.C.; Data Preprocessing, F.G.; Data Analysis, F.G., Writing—original draft, F.G. and A.C.; Writing—review and editing, F.G. and A.C. All authors have read and agreed to the published version of the manuscript.

\section*{Competing interests}
The authors declare no competing interests.

\section*{Ethical approval}
Since the study involves only physicians and no patients, it is not a medical experiment. This work is a part of the broader ADMEDVOICE project, which the Bioethical Commission for Scientific Research has favorably evaluated at the Gdańsk Medical University (Resolution no. KB/508/2023 dated 2023, Sept. 15th). All methods were carried out in accordance with guidelines and regulations from the Bioethical Commission for Scientific Research at the Gdańsk Medical University. Informed consent was obtained from all subjects and/or their legal guardian(s).

\bibliography{references}

\end{document}